\documentclass[runningheads]{llncs}
\usepackage[misc]{ifsym}
\usepackage{graphicx}
\usepackage{amsfonts}
\usepackage{arydshln}
\usepackage{booktabs}
\usepackage{threeparttable}
\usepackage{multirow}
\usepackage[linesnumbered, ruled]{algorithm2e}
\usepackage[colorlinks,
            linkcolor=blue,       
            anchorcolor=blue,  
            citecolor=blue,
            urlcolor=blue
            ]{hyperref}
\usepackage{tikz,xcolor}

\definecolor{lime}{HTML}{A6CE39}
\DeclareRobustCommand{\orcidicon}{%
	\begin{tikzpicture}
	\draw[lime, fill=lime] (0,0) 
	circle [radius=0.16] 
	node[white] {{\fontfamily{qag}\selectfont \tiny ID}};
	\draw[white, fill=white] (-0.0625,0.095) 
	circle [radius=0.007];
	\end{tikzpicture}
	\hspace{-2mm}
}

\foreach \x in {A, ..., Z}{%
	\expandafter\xdef\csname orcid\x\endcsname{\noexpand\href{https://orcid.org/\csname orcidauthor\x\endcsname}{\noexpand\orcidicon}}
}

\begin{document}
\title{Sentence-level Event Detection without Triggers via Prompt Learning and Machine Reading Comprehension}
\titlerunning{Sentence-level Event Detection without Triggers}
%

\author{Tongtao Ling \and Lei Chen\textsuperscript{(\Letter)}\orcidA{}, Huangxu Sheng \and Zicheng Cai \and Hai-Lin Liu\orcidB{}}
%
\authorrunning{T. Ling et al.}
%

\institute{Guangdong University of Technology, Guangzhou, China \\
\email{chenlei3@gdut.edu.cn}}

\maketitle              
\begin{abstract}
The traditional way of sentence-level event detection involves two important subtasks: trigger identification and trigger classifications, where the identified event trigger words are used to classify event types from sentences. However, trigger classification highly depends on abundant annotated trigger words and the accuracy of trigger identification. In a real scenario, annotating trigger words is time-consuming and laborious. For this reason, we propose a trigger-free event detection model, which transforms event detection into a two-tower model based on machine reading comprehension and prompt learning. Compared to existing trigger-based and trigger-free methods, experimental studies on two event detection benchmark datasets (ACE2005 and MAVEN) have shown that the proposed approach can achieve competitive performance.

\end{abstract}

\section{Introduction}\label{sec:intro}

Information extraction (IE) is an important application of Natural Language Processing (NLP). Event detection (ED) is a fundamental part of IE, aiming at identifying trigger words and classifying event types, which could be divided into two sub-tasks: trigger identification and trigger classification~\cite{li2021compact}. For example, consider the following sentence ``\emph{To \textbf{assist} in managing the vessel traffic, Chodkiewicz \textbf{hired} a few sailors, mainly Livonian}''. The trigger words are “assist” and “hired”, the trigger-based event detection model is used to locate the position of the trigger words and classify them into the corresponding event types, \textit{Assistance} and \textit{Employment} respectively.

Nowadays, mainstream research on ED focuses on the trigger-based methods, which means first recognizing triggers and then classifying event types~\cite{chen2015event,sha2018jointly,wang-etal-2019-adversarial-training}. This way transforms the ED task into a multi-stage classification problem, and the result of trigger identification can also affect trigger classification. Therefore, it is crucial to identify trigger words correctly, and this way requires datasets containing multiple annotated trigger words and event types~\cite{lai-etal-2020-extensively}. 
However, it is time-consuming to annotate trigger words in a real scenario, especially in a long sentence. Due to the expensive annotation of the corpus, the application of existing ED approaches is greatly limited. It should be noted that trigger words can provide additional information for trigger classification, but event triggers may not be essential for ED \cite{liu2019event}. 

From a problem-solving perspective, the goal of ED is to categorize the event types, and therefore, triggers can be seen as an intermediate result of this task~\cite{liu2019event}. To alleviate
manual effort, in this paper, we aim to explore how to detect events without triggers.
Event detection can be considered a text classification problem if the event triggers are missing. But three challenges should be solved: (1) \textbf{Multi-label problem}: since a sentence can contain multiple events or no events at all, which is called a multi-label text classification problem in NLP. (2) \textbf{Insufficient event information}: triggers are important and helpful for ED~\cite{chen2015event,zhang-etal-2020-two}. Without trigger words, the ED model may lack sufficient information to detect the event type, and we need to find other ways to enrich the semantic information of the sentence to learn the correlation between the sentence and the event type. (3) \textbf{Imbalance Data Distribution}: the data distribution in the real world is long-tail, which means that most event types have only a small number of instances and many sentences may not have events occurring. The goal of ED is also to evaluate its ability in the long-tail scenario.

\begin{figure}[t]
    \centering    
    \includegraphics[width=1.0\linewidth]{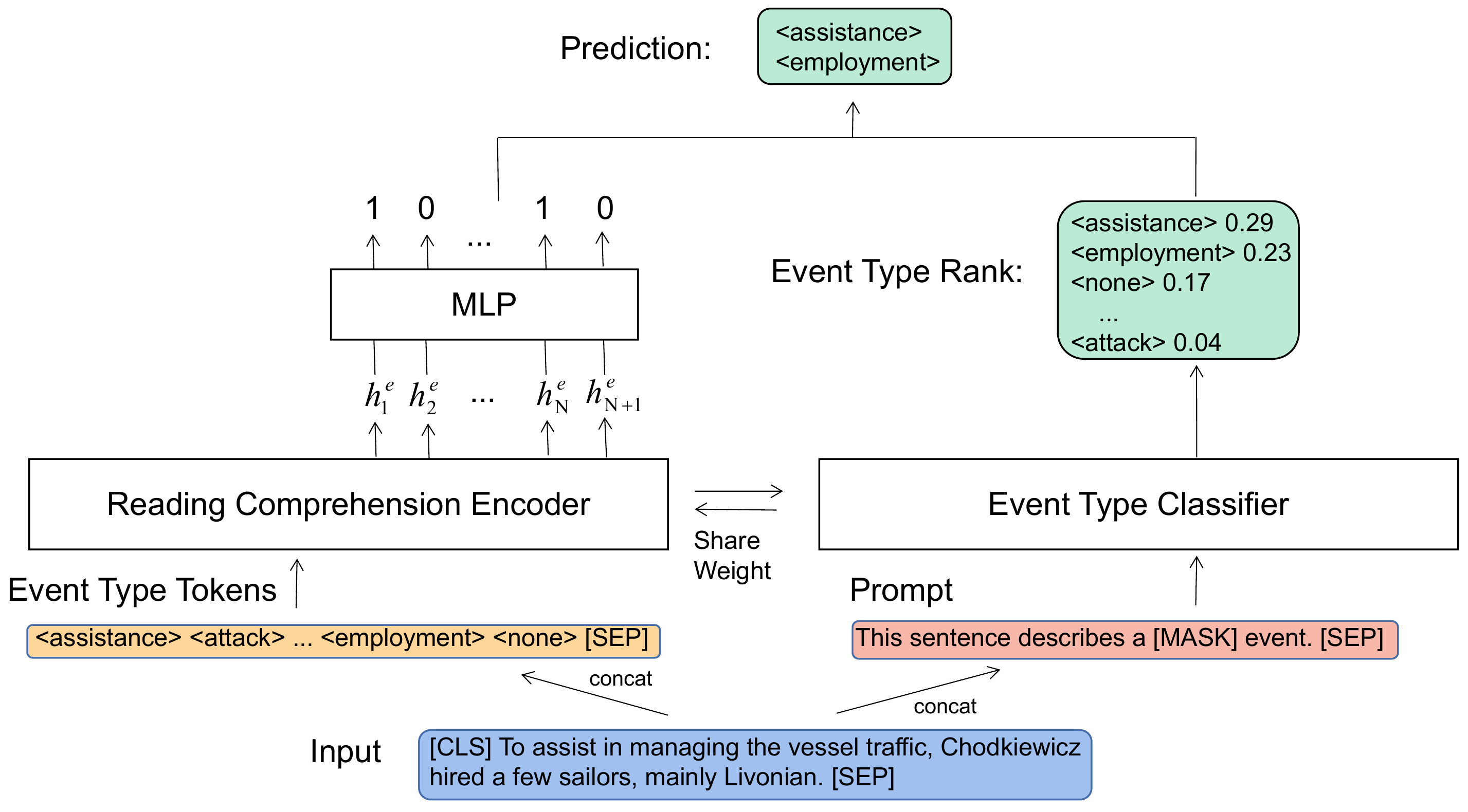}
    \caption{Overview of ED\_PRC. The original input sentence in this example is ``To assist in managing vessel traffic, Chodkiewicz hired a few sailors, mainly Livon''. Token ``[CLS]'' and ``[SEP]'' are special tokens in each sentence before input into BERT.}
    \label{fig:model}
\end{figure}

To address these challenges, we propose a trigger-free method via machine reading comprehension (MRC)~\cite{li2020unified} and prompt learning~\cite{schick-schutze-2021-exploiting} and decompose ED into a two-tower model. Figure \ref{fig:model} illustrates the architecture of our proposed model with two parts: reading comprehension encoder (RCE) and event type classifier (ETC). In the first-tower, we use BERT~\cite{devlin-etal-2019-bert} as the backbone of RCE, and the input sentence concatenates with all event tokens are fed into BERT simultaneously\footnote{For example, we convert event token \textit{employment} to ``$\langle employment \rangle$'' and add it to vocabulary. All events operate like this. In addition, we add a special token ``$\langle none \rangle$'' that no events have occurred.}. Such a way is inspired by the MRC task, extracting event types is formalized as extracting answer position for the given sequence of event type tokens. In other words, the input sentences are deemed as ``Question'' and the sequence of event type tokens deemed as ``Answer''. This way allows BERT to automatically learn semantic relations between the input sentences and event tokens through self-attention mechanism~\cite{vaswani2017attention}. In the second-tower, we use the same backbone of RCE and utilize prompt learning methods to predict event types. Specifically, when adding the prompt “This sentence describes a {\tt [MASK]} event” after the original sentence, this prompt can be viewed as a cloze-style question and the answer is related to the target event type. Therefore, ETC aims to fill the {\tt [MASK]} token and can output the scores for each vocabulary token. We only use event type tokens in vocabulary and predict event types
that score higher than the $\langle none \rangle$ event type. In the inference time, only when these two-tower models predict results are correct can they be used as the final correct answer. In our example from Figure \ref{fig:model}, RCE can predict the answer tokens are $\langle assistance \rangle$ and $\langle employment \rangle$ respectively. In addition, since $\langle assistance \rangle$ and $\langle employment \rangle$ both have higher values than $\langle none \rangle$, we predict \textit{Assistance} and \textit{Employment} as the event type in this sentence.

In summary, we propose a two-tower model to solve the ED task without triggers and call our model \textbf{ED\_PRC}: \textbf{E}vent \textbf{D}etection via \textbf{P}rompt learning and machine \textbf{R}eading \textbf{C}omprehension. The main contributions of our work are: (1) We propose a trigger-free event detection method based on prompt learning and machine reading comprehension that does not require triggers. The machine reading comprehension method can capture the semantic relations between sentence and event tokens. The prompt learning method can evaluate the scores of all event tokens in vocabulary; (2) Our experiments can achieve competitive results compared with other trigger-based methods and outperform other trigger-free methods on two event detection datasets (ACE2005 and MAVEN); (3) Further analysis of attention weight also indicates that our trigger-free model can identify the relation between input sentences and events,  and appropriate prompts in a specific topic can guide pre-trained language models to predict correct events.

\section{Related Work}

\subsection{Sentence-level Event Detection}

Conventional sentence-level event detection models based on pattern matching methods mainly utilize syntax trees or regular expressions~\cite{ahn2006stages}. These pattern-matching methods largely rely on the expression form of text to recognize triggers and classify them into event types in sentences, which fails to learn in-depth features from plain text that contains complex semantic relations. With the rapid development of deep learning, most ED models are based on artificial neural networks such as convolutional neural networks (CNN)~\cite{chen2015event}, recurrent neural network (RNN)~\cite{sha2018jointly}, graph neural network (GNN)~\cite{cui-etal-2020-edge} and transformer network~\cite{yang2019exploring}, and other pre-trained language models~\cite{devlin-etal-2019-bert,wei-etal-2022-desed}. 

\subsection{Machine Reading Comprehension}

Machine reading comprehension (MRC) is a challenging task in natural language processing (NLP). Given a question and a passage, the goal of MRC is to extract answer spans from the passage~\cite{seo2016bidirectional,shen2017reasonet}. In a general way, it is divided into two binary classifiers, predicting the starting and ending spans of the passage. Over the past few years, there has been a trend of transforming event extraction tasks to MRC question answering. For example, Liu et al. \cite{liu-etal-2020-event} explicitly cast event extraction as a MRC task, which transfers event schema into questions and then retrieves answers as results. Zhao et al. \cite{zhao2022trigger} proposes Derangement mechanism on a machine Reading Comprehension (DRC) framework and leverages the self-attention mechanism to absorb semantic relations between context and events.

\subsection{Prompt Learning}

In recent years, prompt-based methods have shown great success in a range of NLP tasks. Compared to the typical model fine-tuning paradigm, prompt-tuning can elicit knowledge from the pre-trained language model (e.g., BERT~\cite{devlin-etal-2019-bert} and GPT3~\cite{brown2020language}) by adding prompts to the raw input, called prompt learning~\cite{schick-schutze-2021-exploiting}. This new paradigm can design various prompts to adapt different downstream tasks (e.g., text classification, relation extraction and text generation), which narrow the gap between the pre-trained tasks and downstream tasks and greatly reduce the training time~\cite{liu2023pre}. Prompt-based learning approaches allow PLMs to have a priori knowledge of a particular downstream task, which contributes to the final performance~\cite{wei2022eliciting}.

\section{Methodology}
\label{sec:method}
In this section, we present the details of the proposed ED\_PRC for sentence-level event detection without triggers.

\subsection{Task Description}

Formally, denote $\mathcal{S}$, $\mathcal{Y}$ as sentence set and event type set, respectively. $\mathcal{S}$ = $\{x_i | i \in [1,M] \}$ contains $M$ sentences, and each sentence $x_i$ in $\mathcal{S}$ is a token sequence $x_i$ = $(w_1,w_2,...,w_L)$ with maximum length $L$. In sentence-level event detection, given a sentence $x_i$ and its ground-truth $y_{i} \in \mathcal{Y}$, $ \mathcal{Y} = \{e_1,e_2,...,e_{N}\}$, we need to detect the corresponding event types for each instance. For sentences where no event occurred, we add a special token ``$\langle None \rangle$'' as their event type. This task can be reformulated as a multi-label classification problem with $N+1$ event types. 

\subsection{Reading Comprehension Encoder}

Inspired by the machine reading comprehension (MRC) task, we utilize BERT as a backbone to design a reading comprehension encoder due to its capability in learning contextual representations of the input sequence. We describe it as follows:
\begin{equation}
   Input =  \textbf{[CLS] Sentence [SEP] Events} 
   \label{eq:eq1}
\end{equation}
where \textbf{Sentence} is the input sentence and \textbf{Events} is the set of event types (also including ``$\langle None \rangle$''). [CLS] and [SEP] are special tokens in BERT. For some event types such as ``Business:Lay off'' fails to map to a single token according to the vocabulary. In this case, we employ an angle bracket around each event type and remove the prefix, e.g., the event type of ``Business:Lay off'' is converted to a lower-case ``$\langle lay\_off\rangle$''. Then, we add $N+1$ event tokens to the vocabulary and randomly initialize its embeddings. We aim to make use of BERT to learn the relation between the input sentence and event types and yield precise event token representations.

After that, we get the hidden-states of the last layer of BERT:
\begin{equation}
   h_{[CLS]}, h_{1}^{w}, ..., h_{L}^{w}, h_{[SEP]}, h_{1}^{e}, ..., h_{N}^{e}, h_{N+1}^{e}  =  BERT(Input)
   \label{eq:eq2}
\end{equation}
where $h_{i}^{w}$ is the hidden state of the $i$-th input token. This setup is close to MRC that chooses the correct option to answer question ``What happened in the sentence?''. Unlike traditional fine-tuning methods that utilize the [CLS] token to complete classification, we use the hidden states of event tokens to predict the probability of each token being the correct answer. The representation of event tokens:
\begin{equation}
    E = h_{1}^{e}, ..., h_{N}^{e},h_{N+1}^{e}
   \label{eq:eq3}
\end{equation}
where $E \in \mathbb{R}^{N \times D}$, $D$ is the vector dimension of BERT. The probability of each event token as follows:
\begin{equation}
    P = softmax(E \cdot W) \in \mathbb{R}^{N \times 2}
   \label{eq:eq4}
\end{equation}
where $W \in \mathbb{R}^{D \times 2} $ is the weight matrix to learn. Each row of $P$ is the probability of the corresponding event type. During training time, we therefore have the following loss for predictions:
\begin{equation}
    \mathcal{L}_{RCE} = CE(P,Y)
   \label{eq:eq5}
\end{equation}
where $Y$ is the ground-truth label of each event token $e_{i}$ being the correct answer.

\subsection{Event Type Classifier}

We describe the implementation of ETC in this subsection. Inspired by the cloze-style prompt learning paradigm for text classification with pre-trained language models, event type classification can be realized by filling the {\tt [MASK]} answer using a prompt function. 

First, the prompt function wraps the input sentence by inserting pieces of natural language text. For prompt function $f_{p}$, as illustrated in Figure~\ref{fig:model}, we use ``{\tt [SENTENCE]} This sentence describes a {\tt [MASK]} event'' as a prompt function for our model. Let $\mathcal{M}$ be pre-trained language model (i.e., BERT), and $\mathbf x$ be the input sentence. The prediction score of each token $v$ in vocabulary being filled in {\tt [MASK]} token can be computed as:
\begin{equation}
  p_{v} = \mathcal{M}({\tt [MASK]} = v| f_{p}(x))
\label{eq:eq6}
\end{equation}

After that, the other key of prompt learning is answer engineering. We aim to construct a mapping function from event token space to event type space. In the first tower (RCE), it learns the relation between the input sentence and event tokens. RCE and ETC share the same weights of BERT. Then, we only select tokens in $\mathcal{Y} = \{e_1,e_2,...,e_{N}\}$ and compute the scores of event tokens:
\begin{equation}
  p_{e} = \sigma( p_{v} |  v \in \mathcal{Y})
\label{eq:eq7}
\end{equation}
where $\sigma(\cdot)$ determines which function to transform the scores into the probability of event tokens, such as \textit{softmax}.

Finally, as shown in Figure \ref{fig:model}, we predict all event tokens that score higher than the ``$\langle None\rangle$'' token as the predicted result. In our example, since both ``$\langle assistance \rangle$'' and ``$\langle employment \rangle$'' have higher scores than ``$\langle None \rangle$'', we predict \textit{Assistance} and \textit{Employment} as target event types.

In the process of training, we calculate two losses due to the problem of imbalance data distribution. The first loss is defined as:
\begin{equation}
  \mathcal{L}_{1} = \frac{1}{|T|} \sum_{t \in T} \log \frac{\exp (\mathcal{M}({\tt [MASK]} = t| f_{p}(x))) } { \sum_{t^{\prime} \in \{ t, \langle none \rangle \} } \exp (\mathcal{M}({\tt [MASK]} = t^{\prime}| f_{p}(x))) }
\label{eq:eq8}
\end{equation}
where $T$ is the set of event tokens that score higher than ``$\langle None\rangle$'' in the sentence. The second loss is defined as follows:
\begin{equation}
  \mathcal{L}_{2} = \log \frac{\exp (\mathcal{M}({\tt [MASK]} = \langle none \rangle | f_{p}(x))) } { \sum_{t^{\prime} \in \{ \langle none \rangle \} \cup \overline{T} } \exp (\mathcal{M}({\tt [MASK]} = t^{\prime}| f_{p}(x))) }
\label{eq:eq9}
\end{equation}
where $\overline{T}$ is the set of event tokens that score lower than ``$\langle None\rangle$'' in the sentence. Note that in Equation \ref{eq:eq8}, we only compare the prediction scores that higher than the ``$\langle None\rangle$'' event token. The reason is that we aim to improve the score of each event token that is higher than ``$\langle None\rangle$''. In Equation \ref{eq:eq9}, we compare to event tokens that lower than the ``$\langle None\rangle$'', which can decrease the score of them. The overall training objective to be minimized is as follows:
\begin{equation}
  \mathcal{L}_{ETC} = \frac{1}{M} \sum_{x \in \mathcal{S}} (\mathcal{L}_{1} + \mathcal{L}_{2})
\label{eq:eq10}
\end{equation}

In the training time, our model can then be trained by minimizing the following loss:

\begin{equation}
  \mathcal{L} = \mathcal{L}_{RCE} + \mathcal{L}_{ETC}
\label{eq:eq11}
\end{equation}

\section{Experiments}
\label{sec:expers}

In this section, we introduce the dataset, evaluation metrics, implementation details, and experimental results.

\subsection{Dataset and Evaluation}

To evaluate the potential of \textbf{ED\_PRC} under different size datasets, we conducted our experiments on two event detection benchmark datasets, ACE2005~\cite{doddington2004automatic} and MAVEN~\cite{wang2020maven}. Details of statistics are available in Table \ref{tab:dataset}.
\begin{itemize}
    \item \textbf{ACE2005} is the most widely used multilingual dataset for event extraction. We use the English version which contains 599 documents and 33 event types. Following previous pre-processing for data split, we adopt two variants: ACE05-E~\cite{wadden-etal-2019-entity} and ACE05-E$^{+}$~\cite{lin-etal-2020-joint}. Compared to ACE05-E, ACE05-E$^{+}$ add pronoun roles and multi-token event triggers. 
    \item \textbf{MAVEN} is a large event detection dataset constructed from Wikipedia and FrameNet, covering 4,480 documents and 168 event types. 
\end{itemize}

For data split and preprocessing, following previous work~\cite{wadden-etal-2019-entity,lin-etal-2020-joint,wang2020maven}, we split 599 documents of ACE2005 into 529/30/40 for train/dev/test set, respectively. Then, we use the same processing that splits 4480 documents of MAVEN into 2913/710/857 for train/dev/test set respectively.  

Following the widely-used metrics for event detection~\cite{chen2015event}, we use precision (P), recall (R) and mirco F1-score (F1) to evaluate results.

\begin{table}[h]
  \centering
  \begin{threeparttable}
  \caption{Dataset statistics of ACE05-E, ACE05-E$^{+}$ and MAVEN.}
    \begin{tabular}{ccccc}
    \toprule
     \textbf{Dataset} & \textbf{Split} & \textbf{\#Sentences} & \textbf{\#Events} & \textbf{\#Documents} \\
    \midrule
        \multirow{3}*{\textbf{ACE05-E}}
        &Train&17,172&4,202&529\\
        &Dev&923&450&30\\
        &Test&832&403&40\\

    \midrule
        \multirow{3}*{\textbf{ACE05-E$^{+}$}}
        &Train&19216&4419&529\\
        &Dev&901&468&30\\
        &Test&676&424&40\\
    \midrule
        \multirow{3}*{\textbf{MAVEN}}
        &Train&32431&73496&2913\\
        &Dev&8042&17726&710\\
        &Test&9400&20389&857\\
    \bottomrule
    \end{tabular}
    \label{tab:dataset}
    \end{threeparttable}
\end{table}

\subsection{Baseline}
We compare our method to baselines with trigger-based and trigger-free methods. For trigger-based methods, we compare with: (1)\textbf{DMCNN}~\cite{chen2015event}, which utilizes a convolutional neural network (CNN) and a dynamic multi-pooling mechanism to learn sentence-level features; (2) \textbf{BiLSTM}~\cite{hochreiter1997long}, which uses bi-directional long short-term memory network (LSTM) to capture the hidden states of triggers and classify them into corresponding event types; (3)\textbf{MOGANDED}~\cite{yan-etal-2019-event}, which proposes multi-order syntactic relations in dependency trees to improve event detection; (4)\textbf{BERT}~\cite{devlin-etal-2019-bert}, fine-tuning BERT on the ED task via a sequence labeling manner; (5)\textbf{DMBERT}~\cite{wang-etal-2019-adversarial-training}, which adopts BERT as backbone and utilizes a dynamic multi-pooling mechanism to aggregate textual features. For trigger-free methods, we compare with: (6)\textbf{TBNNAM}~\cite{liu2019event}, the first work on detecting events without triggers, which uses LSTM and attention mechanisms to detect events; (7)\textbf{TEXT2EVENT}~\cite{lu2021text2event}, proposing a sequence-to-sequence model and extracting events from the text in an end-to-end manner; (8)\textbf{DEGREE}~\cite{hsu-etal-2022-degree}, formulating event detection as a conditional generation problem and extracting final predictions from the generated sentence with a deterministic algorithm.

We re-implemented some trigger-based baselines for comparison, including DMCNN, BiLSTM, MOGANDED, BERT and DMBERT. The other baseline results are from the original paper.

\subsection{Implementation Details}

The proposed model is implemented on the basis of Transformers toolkit~\cite{wolf2020transformers} and PyTorch. We employ \textit{bert-base-uncased}\footnote{\url{https://huggingface.co/bert-base-uncased}} as the backbone and use AdamW as optimizer with a learning rate of 2e-5, max gradient norm of 1.0 and weight decay 5e-5. The maximum sequence length is set to 128 for ACE2005 and 256 for MAVEN. The dropout rate is set to 0.3 and batch size is set to 8. Our model is trained for 10 epochs and chooses the checkpoint with the best validation performance on the development set. We run all experiments on a single Nvidia RTX 3090 GPU. Our code is available at \url{https://github.com/rickltt/event_detection}.

\begin{table}[t]
  \centering
  \begin{threeparttable}
  \caption{Event detection results on both trigger-based and trigger-free methods of the ACE2005 corpora. “-” means not reported in original paper. $\ast$ indicates results cited from the original paper.}
    \begin{tabular}{cccccccc}
    \toprule
    \multirow{2}{*}{\textbf{Category}}&\multirow{2}{*}{\textbf{Models}}&
    \multicolumn{3}{c}{\textbf{ACE05-E}}&\multicolumn{3}{c}{\textbf{ACE05-E}$^{+}$}\cr
    \cmidrule(lr){3-5} \cmidrule(lr){6-8} 
    &&P&R&F-1&P&R&F-1\cr
    \midrule
        \multirow{5}*{\textbf{Trigger-based}}
        &DMCNN&74.3&66.8&70.3 &67.0&73.5&70.1 \\
        &BiLSTM&73.6&72.3&72.9 &73.5&71.3&72.4\\
        &MOGANED&74.6&71.1&72.8 &74.2&72.2&73.2\\
        &BERT&72.5&74.2&73.3 &75.2&72.4&73.8 \\
        &DMBERT&76.4&71.9&74.1&74.9&73.5&74.2\\
    \midrule
        \multirow{4}*{\textbf{Trigger-free}}
        &TBNNAM$^{\ast}$&76.2&64.5&69.9&-&-&- \\
        &TEXT2EVENT$^{\ast}$&69.6&74.4&71.9&71.2&72.5&71.8 \\
        &DEGREE$^{\ast}$&-&-&73.3&-&-&70.9 \\
        &ED\_PRC (Ours)&76.1&71.5&73.7&74.6&73.2&73.9 \\
    \bottomrule
    \end{tabular}
    \label{tab:ace}
    \end{threeparttable}
\end{table}

\subsection{Main Results}

We report main results in Table \ref{tab:ace}. Compared with trigger-free methods, we can find out that our method achieves a much better performance than other trigger-free baselines (TBNNAM, TEXT2EVENT and DEGREE). Obviously, ED\_PRC can achieve improvements of 0.4\% (73.3\% v.s. 73.7\%) F1 score of the best trigger-free baseline (DEGREE) in ACE05-E, and 2.1\% (71.8\% v.s. 73.9\%) F1 score of TEXT2EVENT in ACE05-E$^{+}$. It proves the overall superiority and effectiveness of our model in the absence of triggers. Compared to trigger-based methods, despite the absence of trigger annotations, ED\_PRC can achieve competitive results with other trigger-based baselines, which is only 0.4\% (73.7\% vs. 74.1\%) in ACE05-E and 0.3\% (73.9\% vs. 74.2\%) in ACE05-E$^{+}$ less than the best trigger-based baseline (DMBERT). The result shows that prompt-based method can greatly utilize pre-trained language models to adapt ED task and our MRC module is capable of learning relations between the input text and the target event tokens under low trigger clues scenario. 

To further evaluate the effectiveness of our model on large-scale corpora, we show the result of MAVEN on various trigger-based baselines and our model in Table \ref{maven}. We can see that our model also can achieve competitive performance on various trigger-based baselines, reaching 69.1\% F1 score. Compared with CNN-based (DMCNN), RNN-based (BiLSTM) and GNN-based (MOGANED) method, BERT-based methods (BERT, DMBERT and ED\_PRC) can outperform high improvements, which indicates pre-trained language models can greatly capture contextual representation of input text. However, ED\_PRC can achieve only improvements of 0.1\% (67.2\% v.s. 67.3\%) F1 score on BERT and is 0.8\% (67.3\% v.s. 68.1\%) less than DMBERT. This can be attributed to more triggers and events on MAVEN than that on ACE2005. We conjecture that trigger-based event detection models can greatly outperform trigger-free models when sufficient event information is available. All in all, our ED\_PRC is proven competitive in both ACE2005 dataset and MAVEN dataset.

\section{Analysis}

In this section, we demonstrate further analysis and give an insight into the effectiveness of our method.

\subsection{Effective of Reading Comprehension Encoder}

\begin{figure}[t]
    \centering
    \includegraphics[width=1.0\textwidth]{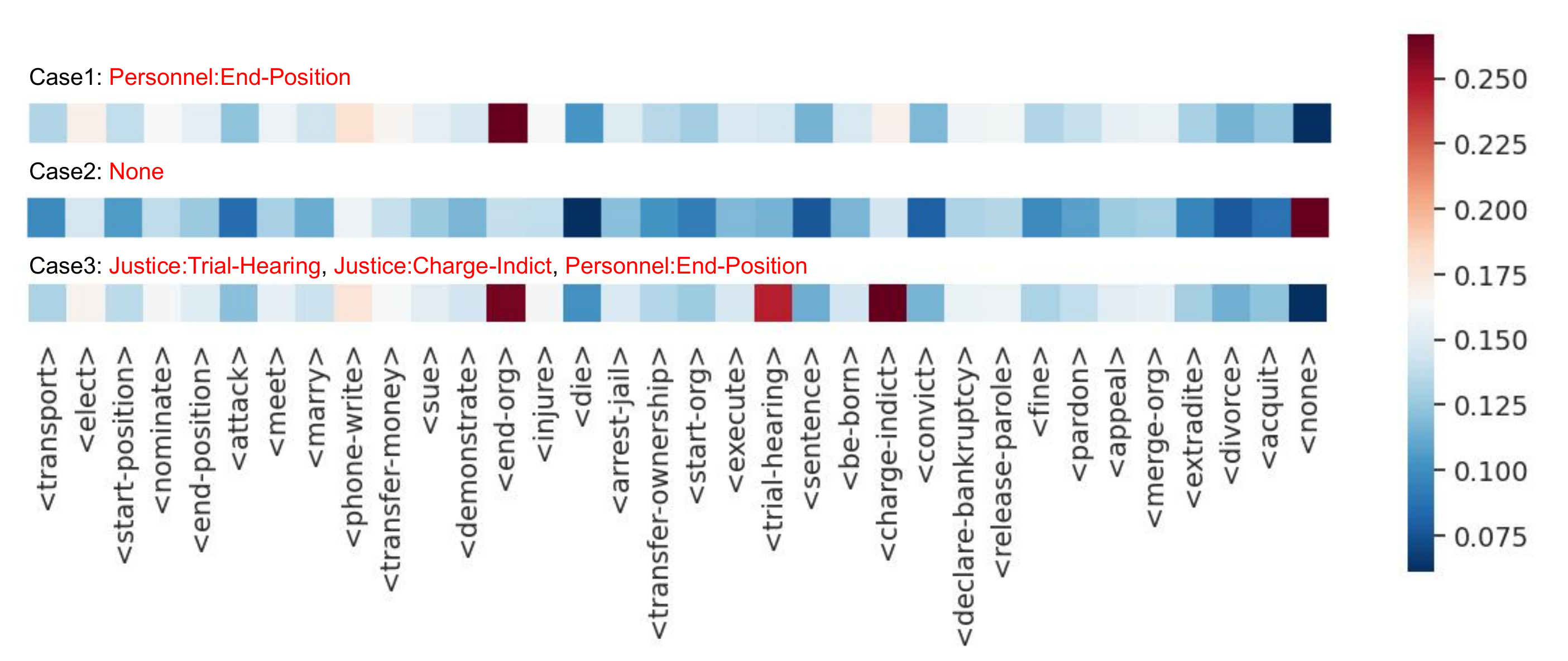}
    \caption{The ACE2005 examples visualization of attention weight in event tokens. We show three cases, the first with only one event, the second with no events and the third with multiple events.}
    \label{fig:attention} 
\end{figure}

Figure \ref{fig:attention} shows a few examples with different target event types and their attention weight visualizations learned by the reading comprehension encoder. In the first case, the target event type is ``Personnel:End-Position'' and our reading comprehension encoder successfully captures this feature by giving ``$\langle end-org \rangle$''
a high attention score. In addition, in the second case, it is a negative sample that no event happened in this sentence and our reading comprehension encoder can correctly give a high attention score for ``$\langle none \rangle$'' and give low attention scores for other event tokens. Moreover, three events occur in the third case, ``Justice:Trial-Hearing'', ``Justice:Charge-Indict'' and ``Personnel:End-Position'', respectively. Our approach can also give high attention scores to ``$\langle trial-hearing \rangle$'', ``$\langle charge-indict \rangle$'' and ``$\langle end-org \rangle$''. We argue that, although triggers are absent, our model can learn the relations between input text and event tokens and assign the ground-truth event tokens with high attention scores.

\begin{table}[t]
	\centering
	\begin{minipage}{0.4\textwidth}
		\centering
		\small
		\makeatletter\def\@captype{table}\makeatother
		\caption{Event detection results on MAVEN corpus.} 
		\label{maven}
        \begin{tabular}{lccc}
        \toprule
        Models&P&R&F-1 \\
        \midrule
        DMCNN&66.5&58.4&62.2 \\
        BiLSTM&64.7&68.2&62.4 \\
        MOGANED&65.9&65.1&65.5 \\
        BERT&64.3&70.5&67.2 \\
        DMBERT&68.9&67.4&68.1 \\
        ED\_RRC (Ours)&66.0&68.7&67.3 \\
        \bottomrule
        \end{tabular}
	\end{minipage}\quad
	\begin{minipage}{0.4\textwidth}
		\centering
		\small
		\makeatletter\def\@captype{table}\makeatother
		\caption{Results on ACE2005 datasets with different prompts.} 
		\label{prompt}
        \begin{tabular}{lccc}
        \toprule
        Models&P&R&F-1 \\
        \midrule
        Prompt\_1&74.2&72.9&73.5 \\
        Prompt\_2&75.6&71.4&73.4 \\
        Prompt\_3&74.7&71.2&72.9 \\
        Prompt\_4&74.1&73.5&73.8 \\
        Soft&73.5&72.7&73.1 \\
        \bottomrule
        \end{tabular}
	\end{minipage}\quad
\end{table}
\subsection{Effective of Different Prompts}

Generally, as the key factor in prompt learning, the prompt can be divided into two categories: hard prompt and soft prompt. The hard prompt is also called a discrete template, which inserts tokens into the original input sentence. Soft prompt is also called continuous template, which is a learnable prompt that does not need any textual templates. To further analyze the influence of different prompts, we design four simple manual prompts (hard prompt) to predict event types: (1) What happened? {\tt [SENTENCE]} This sentence describes a {\tt [MASK]} event; (2) {\tt [SENTENCE]} What event does the previous sentence describe? It was a {\tt [MASK]} event; (3) {\tt [SENTENCE]} It was {\tt [MASK]}; (4) A {\tt [MASK]} event: {\tt [SENTENCE]}. For soft prompt, we insert four trainable tokens into the original sentence, such as ``{\tt [TOKEN]} {\tt [TOKEN]} {\tt [SENTENCE]} {\tt [TOKEN]} {\tt [TOKEN]} {\tt [MASK]}''. The results of our method on ACE2005 are shown in Table \ref{prompt}. 

Prompt\_1 and Prompt\_2 perform similarly, and both of them work better than Prompt\_3. The reason for this may be that Prompt\_3 provides less information and less topic-specific. And both Prompt\_1 and Prompt\_2 add a common phrase ``sentence describe'' and a question to prompt the model to focus on the previous sentence. Unlike previous prompts, Prompt\_4 puts {\tt [MASK]} at the beginning of a sentence, and the result indicates that it might be slightly better to put the {\tt [MASK]} at the end of the sentence. Compared with hard prompt, soft prompt eliminate the need for manual human design and construct trainable tokens that be optimized during training time. The result of soft prompt achieve performance that was fairly close to the hard prompt.

\section{Conclusion}
\label{sec:conl}

In this paper, we transform sentence-level event detection to a two-tower model via prompt learning and machine reading comprehension, which can detect events without trigger words. By using machine reading comprehension framework to formulate a reading comprehension encoder, we can learn the relation between input text and event tokens. Besides, we utilize prompt-based learning methods to construct an event type classifier and final predictions are based on two towers. To make effective use of prompts, we design four manual hard prompts and compare with soft prompt. Experiments and analyses show that ED\_PRC can even achieves competitive performance compared to mainstream approaches using annotated triggers. In the future, we are interested in exploring more event detection methods without triggers by using prompt learning or other techniques.

\subsubsection{Acknowledgements} This work was supported in part by the National Natural Science Foundation of China (62006044, 62172110), in part by the Natural Science Foundation of Guangdong Province (2022A1515010130), and in part by the Programme of Science and Technology of Guangdong Province (2021A0505110004).

%
%
%
\bibliographystyle{splncs04}
\bibliography{reference}
%




\end{document}